\documentclass[letterpaper]{article} 
\usepackage{aaai23}  
\usepackage{times}  
\usepackage{helvet}  
\usepackage{courier}  
\usepackage[hyphens]{url}  
\usepackage{graphicx} 
\usepackage{natbib}  
\usepackage{caption} 
\usepackage{algorithm}
\usepackage{algorithmic}

\usepackage{newfloat}
\usepackage{listings}
\DeclareCaptionStyle{ruled}{labelfont=normalfont,labelsep=colon,strut=off} 
\floatstyle{ruled}
\newfloat{listing}{tb}{lst}{}
\floatname{listing}{Listing}
\usepackage[capitalize]{cleveref}
\usepackage{booktabs}
\usepackage{multirow}
\usepackage{bm}
\usepackage{amssymb}
\usepackage{subfigure}

\title{Referring Expression Comprehension Using Language Adaptive Inference}
\author{
    Wei Su\textsuperscript{\rm 1}, 
    Peihan Miao\textsuperscript{\rm 2}, 
    Huanzhang Dou\textsuperscript{\rm 1}, 
    Yongjian Fu\textsuperscript{\rm 1}, 
    Xi Li\textsuperscript{\rm 1,\rm 3,\rm 4}\thanks{indicates corresponding author.}
}
\affiliations{
    \textsuperscript{\rm 1}College of Computer Science \& Technology, Zhejiang University\\
    \textsuperscript{\rm 2}School of Software Technology, Zhejiang University\\
    \textsuperscript{\rm 3}Shanghai Institute for Advanced Study, Zhejiang University\\
    \textsuperscript{\rm 4}Shanghai AI Laboratory\\
    \{weisuzju, peihan.miao, hzdou, yjfu, xilizju\}@zju.edu.cn

}

\begin{document}

\maketitle

\begin{abstract}
Different from universal object detection, referring expression comprehension (REC) aims to locate specific objects referred to by natural language expressions.
The expression provides high-level concepts of relevant visual and contextual patterns, which vary significantly with different expressions and account for only a few of those encoded in the REC model.
This leads us to a question: \textit{do we really need the entire network with a fixed structure for various referring expressions?}
Ideally, given an expression, only expression-relevant components of the REC model are required. These components should be small in number as each expression only contains very few visual and contextual clues.
This paper explores the adaptation between expressions and REC models for dynamic inference. Concretely, we propose a neat yet efficient framework named Language Adaptive Dynamic Subnets (LADS), which can extract language-adaptive subnets from the REC model conditioned on the referring expressions. By using the compact subnet, the inference can be more economical and efficient. 
Extensive experiments on RefCOCO, RefCOCO+, RefCOCOg, and Referit show that the proposed method achieves faster inference speed and higher accuracy against state-of-the-art approaches.
\end{abstract}

\section{Introduction}
Referring expression comprehension (REC) \cite{transvg,rccf,resc} aims at locating the target object in an image according to a natural language description, which can facilitate not only human-machine interaction in the physical world but also the advancement of tasks such as visual question answering \cite{li2018tell} and image retrieval \cite{salvador2016faster}.
Different from object detection \cite{carion2020end}, which only localizes objects of predefined categories, REC needs to utilize the characteristics of the target object described by referring expressions, including visual patterns (\textit{e.g.} category and attribute) and contextual patterns (\textit{e.g.} relative position and relationship), to locate the most matching region in the image.

\begin{figure}[t]
    \centering
    \includegraphics[width=0.47\textwidth]{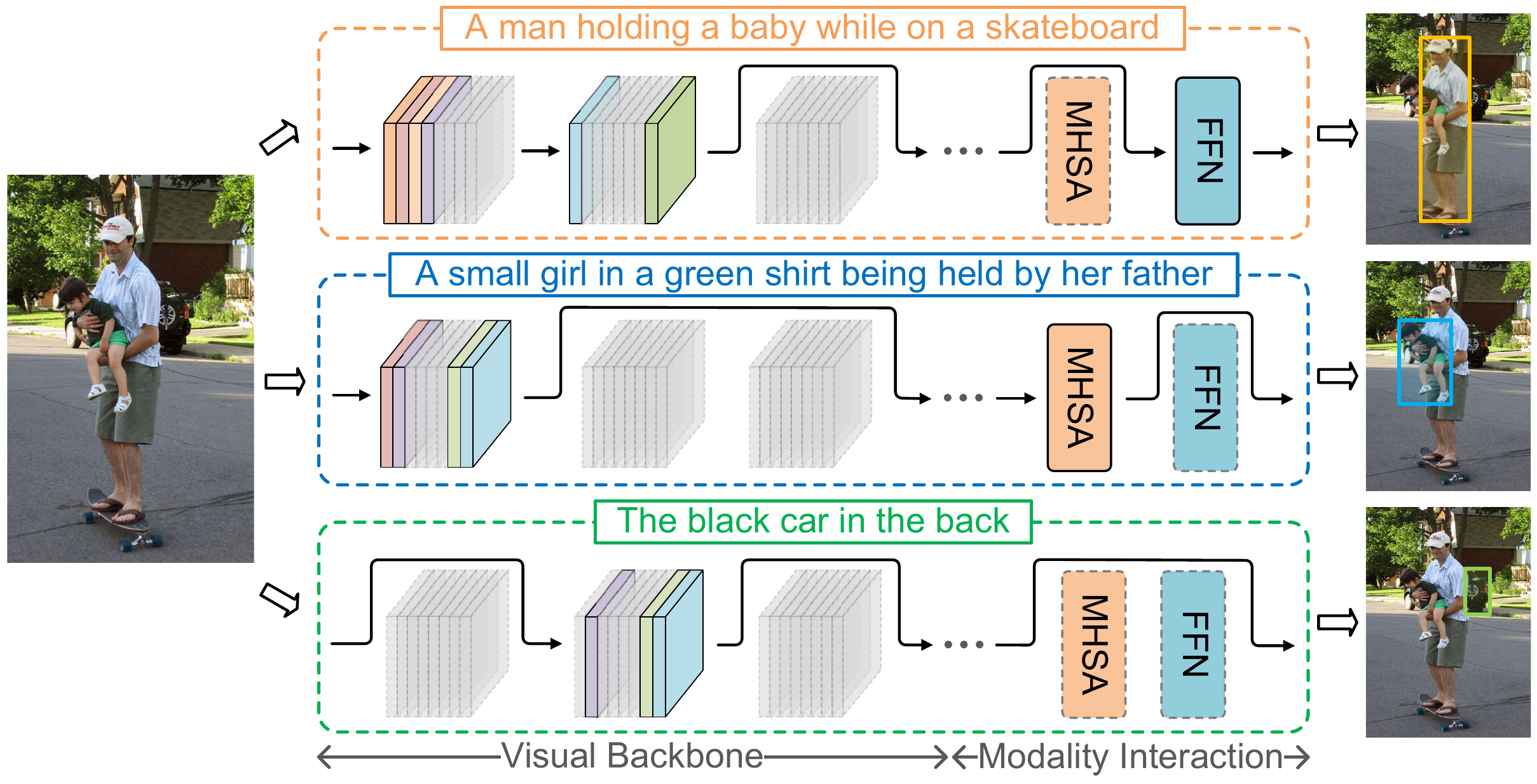}
    \caption{The schematic of the proposed language adaptive dynamic subnet framework (LADS). Given an image and several referring expressions, the LADS framework can select relevant components (\textit{i.e.} filters and layers) from the entire REC network to construct language-adaptive subnets.}
    \label{fig:first_fig}
\end{figure}

As the referring expressions exclusively provide the visual and contextual clues about the referred objects, we overview the REC datasets and summarize two main characteristics of the expressions: (1) \textit{the abundance of semantic information varies widely for different expressions}, and (2) \textit{the visual and contextual patterns required for each expression are only a tiny fraction of the pattern set corresponding to the dataset}.
Taking \cref{fig:first_fig} as an example, the image is assigned three referring expressions, and the complexity of the expressions varies greatly. 
In terms of the first characteristic, the second expression in \cref{fig:first_fig} contains abundant information about the referred object, including three categories ("girl", "father", "shirt"), two attributes (”small”, ”green”), and the relationship ("held").
Conversely, the third expression is less informative, containing only one category ("car") and two attributes ("black", "back"). 
In terms of the second characteristic, there are only 1-3 categories in each expression in \cref{fig:first_fig}, compared to 80 categories in RefCOCO \cite{refcoco} and 238 categories in Referit \cite{referitgame}, which indicates that the category-relevant visual patterns required for each expression are only a tiny proportion of those corresponding to the entire dataset. Similarly, the contextual patterns also account for a very low proportion.

The diversity and low informativeness of referring expressions lead us to an interesting research question: \textbf{Is the entire REC network with a fixed structure necessary for various referring expressions?} 
Intuitively, we need a dynamic inference framework in which we can construct a language-adaptive REC network on the fly, conditioned on different referring expressions. The dynamic networks should be compact and light-weighted since the referring expressions contain very few visual and contextual clues/patterns.
To this end, we utilize the entire REC network as a supernet, which can be viewed as an ensemble of subnets for various referring expressions.
During inference, only the expression-relevant subnets are extracted from the supernet as the specific REC models, and the expression-irrelevant subnets are removed, which is illustrated in \cref{fig:first_fig}.

Concretely, we propose a neat yet efficient framework named Language Adaptive Dynamic Subnets (LADS), which can adaptively extract REC subnets from the supernet conditioned on different referring expressions.
To obtain more flexible and compact subnets, we also propose to select layers and filters from the REC supernet jointly, and apply mutual information to constrain the alignment of subnets and referring expressions. 
Specifically, the linguistic feature of each referring expression is extracted first and then mapped to binary selection gates.
The selection gates are used to pick out layers and filters from the REC supernet, and construct a compact subnet dedicated to the specific referring expression.
It should be noted that the subnets maybe cohere with each other in the initial REC supernet.
During training, LADS tries to strengthen the expression-relevant subnets and mitigate interference between subnets selected for different expressions.
After training, the subnets can be extracted directly for inference without retraining, just like the once-for-all network \cite{ofa}.
Extensive experiments on RefCOCO \cite{refcoco}, RefCOCO+ \cite{refcoco}, RefCOCOg \cite{refcocog}, and Referit \cite{referitgame} show that the proposed framework achieves faster inference speed and higher accuracy compared to state-of-the-art methods, which shows the effectiveness of the LADS framework and its prospects in more compact and real-time REC inference.

The main contributions are summarized as follows:
\begin{itemize}
    \item  To achieve efficient dynamic reasoning in REC, we propose the Language Adaptive Dynamic Subnets (LADS) framework, which can adaptively select compact REC subnets conditioned on the referring expressions.
    \item To obtain flexible and compact subnets, we propose to select layers and filters from the REC supernet jointly for more diverse subnets in depth and width, and apply mutual information to constrain the alignment of subnets and referring expressions for more compact subnets. 
    \item Experiments on four representative datasets show that the proposed method achieves superior inference speed and accuracy compared to state-of-the-art methods, with the language-adaptive compact REC subnets.
\end{itemize}

\section{Related Work}
\subsection{Referring Expression Comprehension}
Most conventional REC methods consist of two stages \cite{mattnet,refnms,hong2019learning,liu2019learning}.
In the first stage, the candidate regions of the input image are obtained by pre-trained object detectors \cite{ren2015faster,he2017mask}.
In the second stage, given a referring expression, the best matching candidate region is chosen as the prediction.
Most two-stage methods improve overall performance through the second stage.
They exploit linguistic and visual features for contextual information mining, and additionally leverage attributes \cite{liu2017referring,mattnet}, object relationships \cite{wang2019neighbourhood,yang2019cross}, phrase co-occurrence \cite{bajaj2019g3raphground}, \emph{etc.} to improve the performance.
Despite the remarkable success, the overall performance of the two-stage methods is limited by the accuracy and speed of candidate region generation in the first stage.
Recently, one-stage methods \cite{rccf,resc,transvg,zhou2021real,mcn} have proliferated and achieved excellent performance.
In the one-stage paradigm, the network densely fuses linguistic and visual features, and directly outputs the target box \cite{transvg,rccf,resc}, which can get rid of the computation-intensive candidate region generation, and the matching process of candidate regions and referring expression in the two-stage paradigm.

\subsection{Conditional Computation}
Conditional computation (also known as dynamic execution) in computer vision aims to dynamically select a subset of convolutional neural networks (CNN) to execute conditioned on the input images. 
Compared to the fixed inference network, it is more computation-efficient because only a part of the CNN is used.
Lots of works related to conditional computation have been studied.
According to the granularity of selected modules, these works can be divided into layer selection \cite{veit2018convolutional,wang2018skipnet}, filter selection \cite{chen2019you,bejnordi2019batch,chen2019self,gao2018dynamic,herrmann2020channel} and layer-filter joint selection \cite{xia2021fully}. 
Unlike the above methods, which make selection decisions based on the input images and gradually choose/skip layers or filters in the feed-forward inference, the proposed method selects layers and filters to construct the language-adaptive subnet based on a given referring expression before the feed-forward process.

\begin{figure*}[ht]
    \centering
    \includegraphics[width=\textwidth]{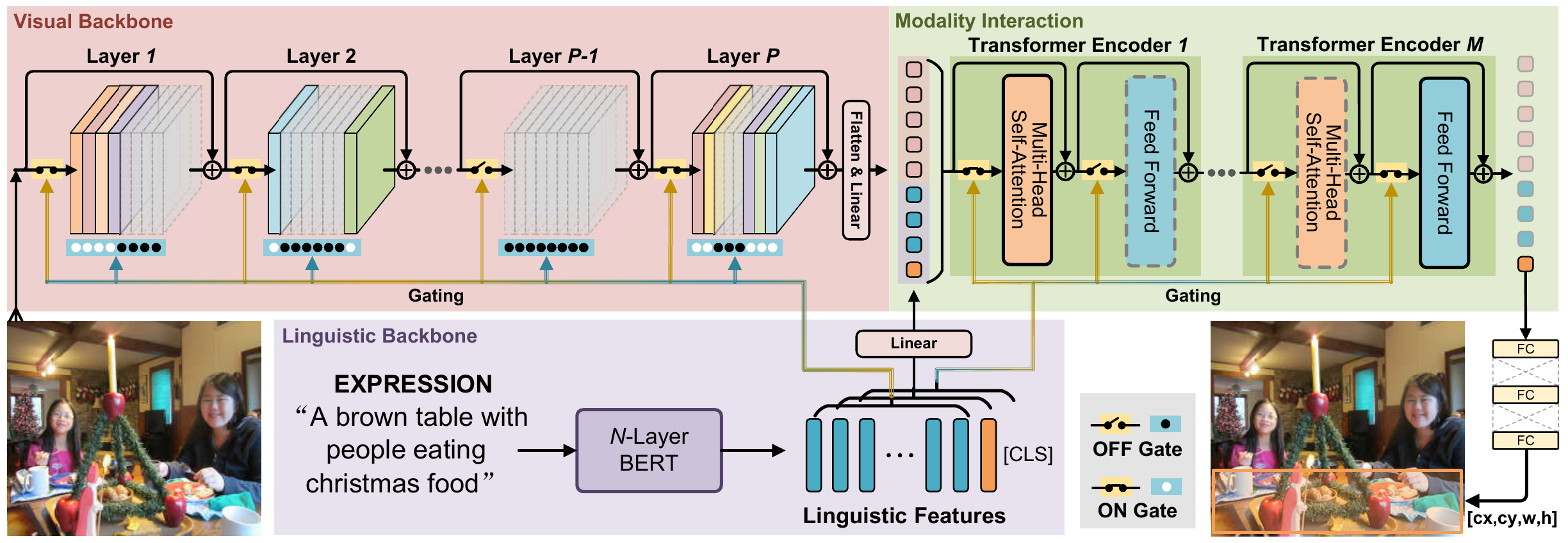}
    \caption{The overall architecture of our proposed LADS framework, which contains a REC supernet and a Gating Network. The REC supernet is composed of three components: (1) Visual Backbone, which extracts and flattens visual features from the input image, (2) Linguistic Backbone, which tokenizes referring expressions and extracts linguistic features with a prepended [CLS] token, and (3) Modality Interaction Module, which fuses the two modal features and regresses the coordinates of the referred objects. The Gating Network is the core module of the LADS framework, which takes linguistic features as input and extracts compact expression-adaptive subnets from the REC supernet by generating binary gates for the layers and filters.}
    \label{fig:main}
\end{figure*}
\section{Approach} 
In this section, we first describe the Language Adaptive Dynamic Subnets (LADS) framework. 
Then we introduce the expression-adaptive subnet selection based on the Gating Network, including the relevance score estimation and binary gate generation.
After that, the mutual information between expressions and subnets is demonstrated.
Finally, the training objectives used for the framework are described.

\subsection{Framework}
\label{sec:framework}
To achieve efficient reasoning while maintaining high accuracy, we propose the Language Adaptive Dynamic Subnets (LADS) framework, as illustrated in \cref{fig:main}.
The LADS framework consists of a REC supernet and a Gating Network.
During inference, the Gating Network first extracts an expression-adaptive subnet.
The subnet is then used for referring grounding.

The REC supernet contains three components, \textit{i.e.} linguistic backbone, visual backbone, and modality interaction module.
Given a referring expression, the $N$-layer BERT-based linguistic backbone \cite{bert} tokenizes the expressions, prepends a [CLS] token, and extracts $d_l$-dimension linguistic features $F_l \in \mathbb{R}^{L \times d_l}$ with $L$ tokens.
Next, given an image, the visual backbone extracts $C$-dimension visual features $F_v \in \mathbb{R}^{C \times H \times W}$ with height $H$ and width $W$.
Then, the visual features $F_v$ are flattened along the spatial dimension to match the valid input dimension of the modality interaction module and added with the sinusoidal positional embeddings to retain the spatial location information.
After that, the linearly projected $F_v$ and $F_l$ are concatenated and fed to the modality interaction module with $M$ transformer encoder layers for cross-modal interaction. 
Finally, we pass the fused features corresponding to the [CLS] token to a 3-layer fully connected (FC) layers, followed by the $Sigmoid$ function to predict the referred bounding box $b=\lbrack cx,cy,w,h \rbrack$, where $(cx,cy)$ and $(w,h)$ denote the center position and size, respectively.

\begin{figure}[t]
    \centering
    \includegraphics[width=0.47\textwidth]{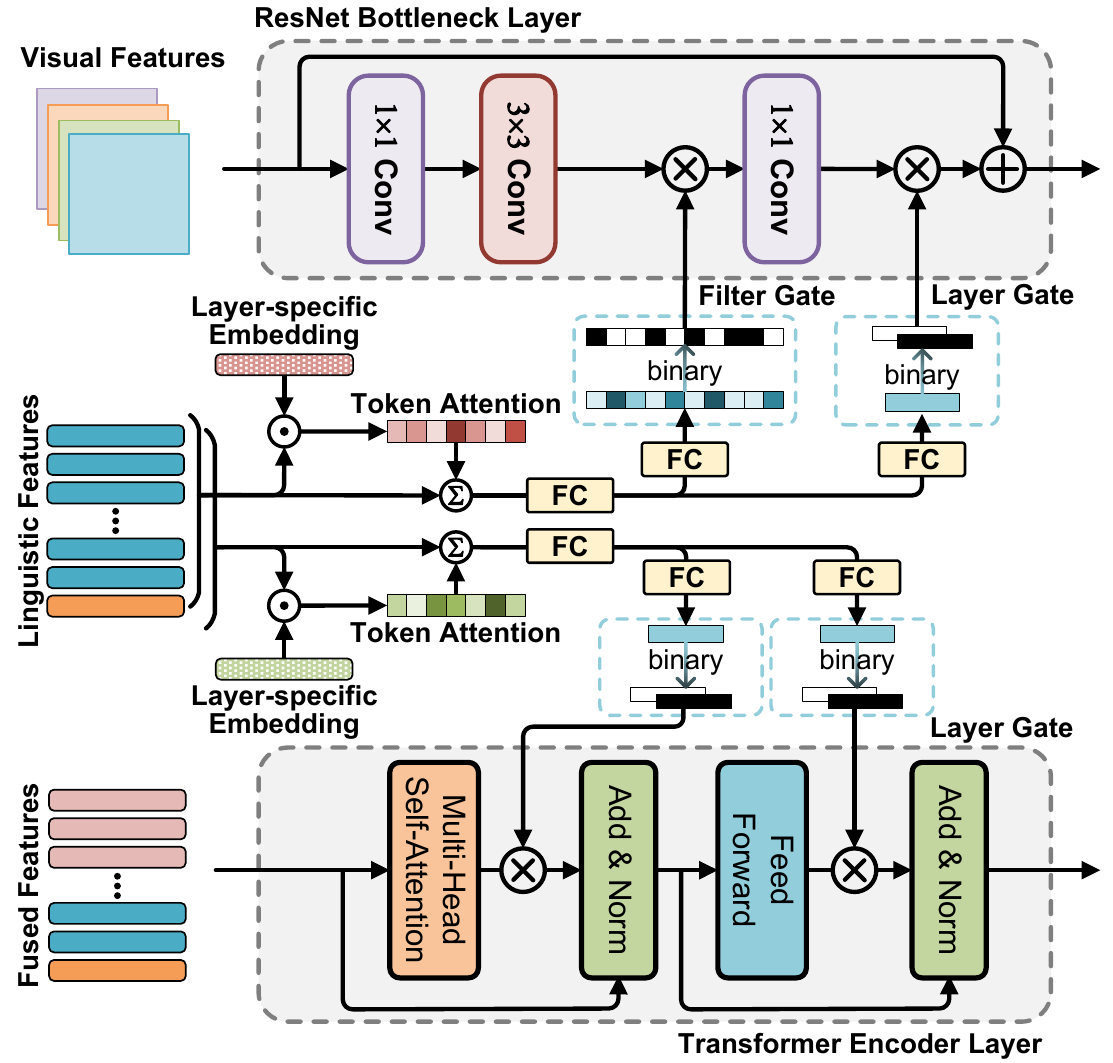}
    \caption{The detailed architecture for gate generation with linguistic features as input. The upper part of this figure shows the gate generation for one ResNet bottleneck layer in the visual backbone. The lower part shows the gate generation for one transformer encoder layer in the modality interaction module.}
    \label{fig:subnet_selection}
\end{figure}

The Gating Network, as the core module of the LADS framework, generates expression-adaptive binary gates based on the linguistic features $F_l$.
Using these binary gates, a compact subnet dedicated to that expression can be extracted from the REC supernet.
To obtain more flexible and compact subnets, we implement bi-level selection, \textit{i.e.} layers and filters. 
Concretely, the residual layers and convolution filters are selected in the visual backbone, and the multi-head self-attention layers and feed-forward layers are selected in the modality interaction module.
The detailed structure of Gating Network is illustrated in \cref{fig:subnet_selection}. 
The Gating Network can be functionally divided into two stages, namely Relevance Score Estimation and Binary Gate Generation, which will be introduced in detail next.

\subsection{Relevance Score Estimation}
\label{sec:relevance_score_estimation}
Different from the previous conditional computation methods, such as GaterNet \cite{chen2019you} and AIG \cite{veit2018convolutional}, which calculate the relevance score with visual features as input, in our LADS framework, the relevance scores are calculated directly according to the linguistic features. 
The calculation can be divided into two steps: (1) pooling the linguistic features for each candidate layer, and (2) mapping the pooled features to relevance scores.

Although there are many token feature pooling methods, such as mean pooling or using [CLS] features, they may not be suitable for our framework because different candidate layers may prefer different tokens.
Therefore, we propose an attention-based parametric pooling method. 
Concretely, we assign a learnable layer-specific embedding $e_i \in \mathbb{R}^{d_l}$ to each candidate layer $i$. 
The token-wise attention weights $\alpha_i \in [0,1]^{L}$ can be derived by calculating the inner product of $e_i$ and $F_l$, followed by $Softmax$ normalization. 
Then, the pooled feature $h_i \in \mathbb{R}^{d_l}$ is set to the weighted sum of $\alpha_i$ and $F_l$. 
The detailed calculation can be denoted as: 
\begin{equation}
    \alpha_i = Softmax([e_i\cdot F_l^1, e_i\cdot F_l^2, \cdots, e_i\cdot F_l^L])
    \label{equ:token_attn}
\end{equation}
\begin{equation}
    h_i = \sum_{j=1}^{L}\alpha_i^j F_l^j
    \label{equ:inner_feat}
\end{equation}

Finally, we utilize two fully-connected layers (FC) to calculate the relevance score $r_L^i$ and $r_C^i$ for the $i$-th layer and filters separately, which can be indicated as:
\begin{equation}
    r_L^i = W_2^i\delta(W_1^i h_i),\ r_C^i = W_3^i\delta(W_1^i h_i)
    \label{equ:relevance_score}
\end{equation}
where shared $W_1^i \in \mathbb{R}^{d_l \times d_h}$ is firstly used to reducing the dimension to $d_h$. $W_2^i\in \mathbb{R}^{1\times d_h}$ and $W_3^i\in \mathbb{R}^{c_i\times d_h}$ are used to calculate the relevance scores, and $c_i$ is the number of filters. $\delta$ refers to the $GeLU$ activation function. Since the multi-head self-attention layer and feed-forward layer have the residual connection, the relevance scores are calculated the same as $r_L^i$.
The Gating Network only contains 0.6M parameters, which is much less than the parameter amount of the whole model.

\subsection{Binary Gate Generation}
\label{sec:binary_gate_generation}
After estimating the relevance between layers and linguistic features, the binary gates are calculated for extracting expression-adaptive subnets from the REC supernet. 

To mitigate model collapse \cite{veit2018convolutional} and maintain differentiability, we leverage the Gumbel-Softmax trick \cite{jang2016categorical, maddison2016concrete}, and simplify it by only calculating the logits that gates are turned on, considering the particularity of the binary decision. 
Correspondingly, we add Logistic noises to the relevance scores instead of the Gumbel noises. 
In addition, inspired by the Improved Semantic Hashing \cite{kaiser2018discrete}, we randomly use the soft and hard gates during training to mitigate the gradient mismatch caused by the straight-through estimation of the hard gates. 
Taking the layer gate $g_L^i \in \mathbb{R}$ of the $i$-th layer as an example, which is calculated as follows:
\begin{equation}
    g_s^i = Sigmoid(\hat{r}_L^i),\ g_h^i = \mathcal{I}(\hat{r}_L^i \ge 0)
    \label{equ:gate_soft_hard}
\end{equation}
\begin{equation}
    g_L^i = \mathcal{I}(n_s \ge 0.5) \cdot g_s^i + \mathcal{I}(n_s < 0.5) \cdot g_h^i
    \label{equ:gate_mix}
\end{equation}
where $\hat{r}_L^i=r_L^i+\epsilon$ is the noisy version of $r_L^i$, and $\epsilon \sim Logistic(0,1)$ represents Logistic noise, which is the difference of two Gumbel i.i.d noises.
$g_s^i \in \left[ 0,1\right]$ and $g_h^i \in \left\{ 0,1 \right\} $ indicate the soft and hard gates, respectively. $\mathcal{I}(\cdot)$ is the indicative function.
$n_s \sim Uniform(0,1)$ is used to randomly select the hard or soft gates in each training batch with a probability $0.5$. The channel gates $g_C^i \in \mathbb{R}^{c_i}$ can be calculated in the same way by replacing $r_L^i$ with $r_C^i$.

The bottleneck layers in the visual backbone, and multi-head self-attention layers and feed-forward layers in the modality interaction module are all residual structures and can be represented as $x_{i+1}=F(x_i)+x_i$. $F(\cdot)$ and $x_i$ denote the residual function and input feature, respectively. 
The layer selection can be achieved by multiplying $F(x_i)$ by layer gate $g_L^i$:
\begin{equation}
    x_{i+1}=F(x_i) \cdot g_L^i+x_i
\end{equation}
Similarly, the filters in the visual backbone can be selected by multiplying the visual feature $y_i$ by filter gates $g_C^i$:
\begin{equation}
    y_i'=y_i \cdot g_C^i
\end{equation}
where $y_i'$ represents the modified visual feature after filter selection. The value of one entire channel is set to $0$ when the corresponding filter gate is $0$.

\begin{table*}[ht]
  \small
  \centering
    \begin{tabular}{c|c|ccc|ccc|cc|c|c}
        \toprule
        \multirow{2}[2]{*}{Method} & Visual & \multicolumn{3}{c|}{RefCOCO} & \multicolumn{3}{c|}{RefCOCO+} & \multicolumn{2}{c|}{RefCOCOg} & Referit & Time \\
              & Backbone & val   & testA & testB & val   & testA & testB & val & test & test  & (ms) \\
        \midrule
        MAttNet \cite{mattnet} & RN101 & 76.40  & 80.43  & 69.28  & 64.93  & 70.26  & 56.00  & 66.67  & 67.01  & -     & 325 \\
        FAOA \cite{faoa} & DN53  & 72.54  & 74.35  & 68.50  & 56.81  & 60.23  & 49.60  & 61.33  & 60.36  & 60.67  & 39 \\
        RCCF \cite{rccf} & DLA34 & -     & 81.06  & 71.85  & -     & 70.35  & 56.32  & -     & 67.01  & 63.79  & 25 \\
        MCN  \cite{mcn} & DN53  & 80.08  & 82.29  & 74.98  & 67.16  & 72.86  & 57.31  & 66.46  & 66.01  & -     & 51 \\
        ReSC \cite{resc} & DN53  & 77.63  & 80.45  & 72.30  & 63.59  & 68.36  & 56.81  & 67.30  & 67.20  & 64.60  & 53 \\
        TransVG \cite{transvg} & RN50  & 80.32  & 82.67  & 78.12  & 63.50  & 68.15  & 55.63  & 67.66  & 67.44  & 69.76  & 41 \\
        Ref-NMS \cite{refnms} & RN101 & 80.70  & 84.00  & 76.04  & 68.25  & 73.68  & 59.42  & 70.55  & 70.62  & -     & - \\
        LADS (ours) & RN50  & \textbf{82.85} & \textbf{86.67} & \textbf{78.57} & \textbf{71.16} & \textbf{77.64} & \textbf{59.82} & \textbf{71.56} & \textbf{71.66} & \textbf{71.08} & \textbf{20} \\
        \midrule
        ViLBERT \cite{vilbert} & RN101 & -     & -     & -     & 72.34  & 78.52  & 62.61  & -     & -     & -     & 400 \\
        UNITER\_L \cite{uniter} & RN101 & 81.41  & 87.04  & 74.17  & 75.90  & 81.45  & 66.70  & 74.86  & 75.77  & -     & 416 \\
        VILLA\_L \cite{villa} & RN101 & 82.39  & 87.48  & 74.84  & 76.17  & 81.54  & 66.84  & 76.18  & 76.71  & -     & 417 \\
        MDETR \cite{mdetr} & RN101 & 86.75  & 89.58  & 81.41  & 79.52  & 84.09  & 70.62  & 81.64  & 80.89  & -     & 65 \\
        LADS (ours) & RN50  & \textbf{87.80} & \textbf{91.23} & \textbf{84.03} & \textbf{79.65} & \textbf{84.86} & \textbf{71.97} & \textbf{82.67} & \textbf{81.96} & \textbf{78.82} & \textbf{18} \\
        \bottomrule
    \end{tabular}
  \caption{Comparisons with state-of-the-art methods on the RefCOCO, RefCOCO+, RefCOCOg, and Referit datasets. RN101, RN50, and DN53 are shorthand for the ResNet101, ResNet50, and DarkNet53, respectively. The inference time is tested on 1080 Ti GPU and averaged over all referring expressions.}
  \label{tab:main_result}
\end{table*}

\subsection{Mutual Information between Expressions and Subnets}
\label{sec:mutual_information}
To encourage compact subnets and enforce the alignment between the expressions and corresponding subnets, we calculate and maximize the mutual information between the representations of expression and subnet. 
Considering that the expression can be represented by the linguistic features $F_l$, and the subnet is uniquely determined by the binary gate set $G=\{g_L, g_C\}$, the mutual information can thus be represented as:
\begin{equation}
    I(G;F_l) = \mathcal{H}(G) - \mathcal{H}(G|F_l)
    \label{equ:mutual_information}
\end{equation} 
where $\mathcal{H}(p)=-\sum_{i=1}^Np_ilog(p_i)$ denotes the entropy of one gate. 
In our LADS framework, considering that each gate contains only two states (\textit{i.e.} "ON" and "OFF"), the gate entropy can be simplified as $\mathcal{H}(p)=-plog(p)-(1-p)log(1-p)$, where $p$ specifies the probability that the gate is "ON". 
We set the conditional probability $p(g|F_l)=g_s$, where $g_s$ is the soft gate mentioned in \cref{equ:gate_soft_hard}, and the gate probability is set to $p(g)\approx\frac{1}{B}\sum_{i=1}^Bp(g|F_l^i)$, which is evaluated in each training mini-batch.

By maximizing the mutual information $I(G;F_l)$, the two terms, \textit{i.e.} gate entropy $\mathcal{H}(G)$ and conditional entropy $\mathcal{H}(G|F_l)$ are maximized and minimized, respectively. 
Apart from enforcing the alignment between expressions and subnets, there are two additional considerations for adopting $I(G;F_l)$: 
(1) Maximizing $\mathcal{H}(G)$ forces the framework to turn on/off the gates evenly, which can avoid the self-reinforcing problem \cite{shazeer2017outrageously} in the training process. The self-reinforcing usually exists in the conditional computation methods, and it can damage the supernet by excluding some layers or filters with zero selection probability in the early training process. 
Benefiting from the $\mathcal{H}(G)$, the self-reinforcing problem can be overcome in our LADS framework by preventing the gates from always turning on or off, and the actual network capacity and accuracy can be kept without layer/filter dropping.
(2) Minimizing $\mathcal{H}(G|F_l)$ constraints the framework to generate unambiguous gates for each sample conditioned on the specific referring expression. 
In addition, it can also push the soft gates to hard gates in \cref{equ:gate_mix} by optimizing the gate probability to $1$ or $0$, and the gap between training and evaluation is narrowed.

\subsection{Training Objectives}
The LADS framework is trained end-to-end, and similar to DETR \cite{carion2020end} and TransVG \cite{transvg}, the L1 loss and Generalized IoU (GIoU) loss are used between the predicted referring box $b=(cx,cy,w,h)$ and ground truth bounding box $\hat{b}=(\hat{cx},\hat{cy},\hat{w},\hat{h})$. The negative mutual information $-I(G;F_l)$ in \cref{equ:mutual_information} is also included to constrain the alignment between the referring expressions and the corresponding REC subnets. The total training loss can be summarized as: 
\begin{equation}
    \mathcal{L}_{total} = \mathcal{L}_{L1}(b,\hat{b}) + \lambda_{giou} \mathcal{L}_{giou}(b,\hat{b}) - \lambda_{mi} I(G;F_l)
\end{equation}
The $\lambda_{giou}$ and $\lambda_{mi}$ are coefficients for the GIoU loss and negative mutual information loss, which are set to $1.0$ and $0.1$ in the experiments, respectively.

\section{Experiments}
In this section, we conduct experiments on the proposed Language Adaptive Dynamic Subnets (LADS) framework to evaluate the accuracy and efficiency on REC datasets. The dynamicity of the REC subnets and the alignment between subnets and expressions are also evaluated.

\subsection{Experimental Setting}
\subsubsection{Datasets}
There are five REC datasets used in the experiments, including RefCOCO \cite{refcoco}, RefCOCO+ \cite{refcoco}, RefCOCOg \cite{refcocog}, Referit \cite{referitgame} and large-scale pre-training dataset.
RefCOCO and RefCOCO+, which are officially split into train, val, testA, and testB sets, have 19,994 images with 142,210 referring expressions and 19,992 images with 141,564 referring expressions, respectively.
RefCOCOg \cite{umd} has 25,799 images with 95,010 referring expressions, which is officially split into train, val, and test sets.
Referit has 20,000 images collected from SAIAPR-12 \cite{saiapr12}, which is split into train and test sets.
The large-scale pre-training dataset has 174k images with approximately 6.1M distinct referring expressions, which contains the train sets of RefCOCO/+/g, Referit, VG regions \cite{vg}, and Flickr entities \cite{flickr30k}.

\subsubsection{Evaluation Metric}
We use \emph{Prec}@0.5 evaluation protocol to evaluate the accuracy. Given a referring expression, a predicted region is considered correct if its intersection-over-union (IoU) with the ground-truth bounding box is greater than 0.5.
In addition, we also report the average time taken for a complete inference of the LADS framework.

\begin{figure*}
      \centering
      \includegraphics[width=\textwidth]{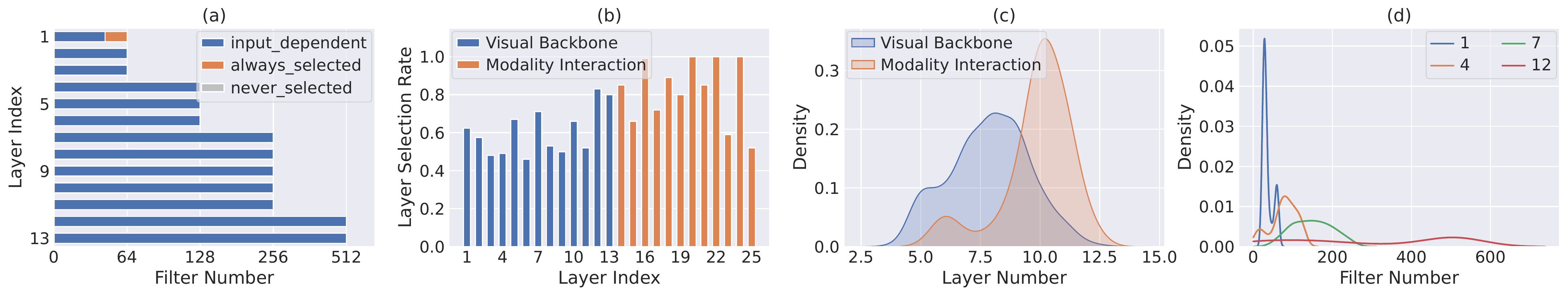}
      \caption{Dynamicity of the REC subnets. From left to right: (a) dynamicity of the filter selection, (b) dynamicity of the layer selection, (c) distribution over the number of selected layers, and (d) distribution over the number of selected filters.}
      \label{fig:filter_layer_disp}
\end{figure*}

\subsection{Implementation Details}
\subsubsection{Training}
All models are trained on the NVIDIA A100 GPU with CUDA 11.4.
For the visual backbone, we use ResNet50 \cite{resnet} pre-trained on MSCOCO \cite{mscoco}, where overlapping images in the val/test sets of the corresponding datasets are excluded.
For the linguistic backbone, we use the first six layers of BERT \cite{bert} provided by HuggingFace \cite{huggingface}.
The rest of the model is initialized using Xavier initialization \cite{glorot2010understanding}.
The input images are resized to $512 \times 512$, and the max expression length is 40.
All models are end-to-end optimized by AdamW \cite{adamw} optimizer with weight decay of 1e-4. The initial learning rate of visual backbone and linguistic backbone is 1e-5, and the initial learning rate of the rest is 1e-4.
We train for 120 epochs with a batch size of 256, where the learning rate is reduced by 10 after 90 epochs.
In large-scale pre-training and fine-tuning, we train for 40 and 20 epochs with batch sizes of 512 and 256, where the learning rate is reduced by 10 after 30 and 10 epochs, respectively.
Following the common practice in \cite{transvg,resc}, we perform data augmentation at the training stage, including random resize, random crop, and horizontal flip.

\subsubsection{Evaluation and Inference}
In the evaluation and inference stage, there are two differences from training. 
First, the Logistic noise $\epsilon$ in \cref{equ:gate_soft_hard} is set to $0$, and there is no randomness in the model. 
Second, the hard gates $g_h^i$ in \cref{equ:gate_mix} are always used, which means that the layers and filters are directly removed when $g_h^i=0$ and the rest components with $g_h^i=1$ make up the expression-adaptive subnets.  

\begin{figure}
    \centering
    \includegraphics[width=0.45\textwidth]{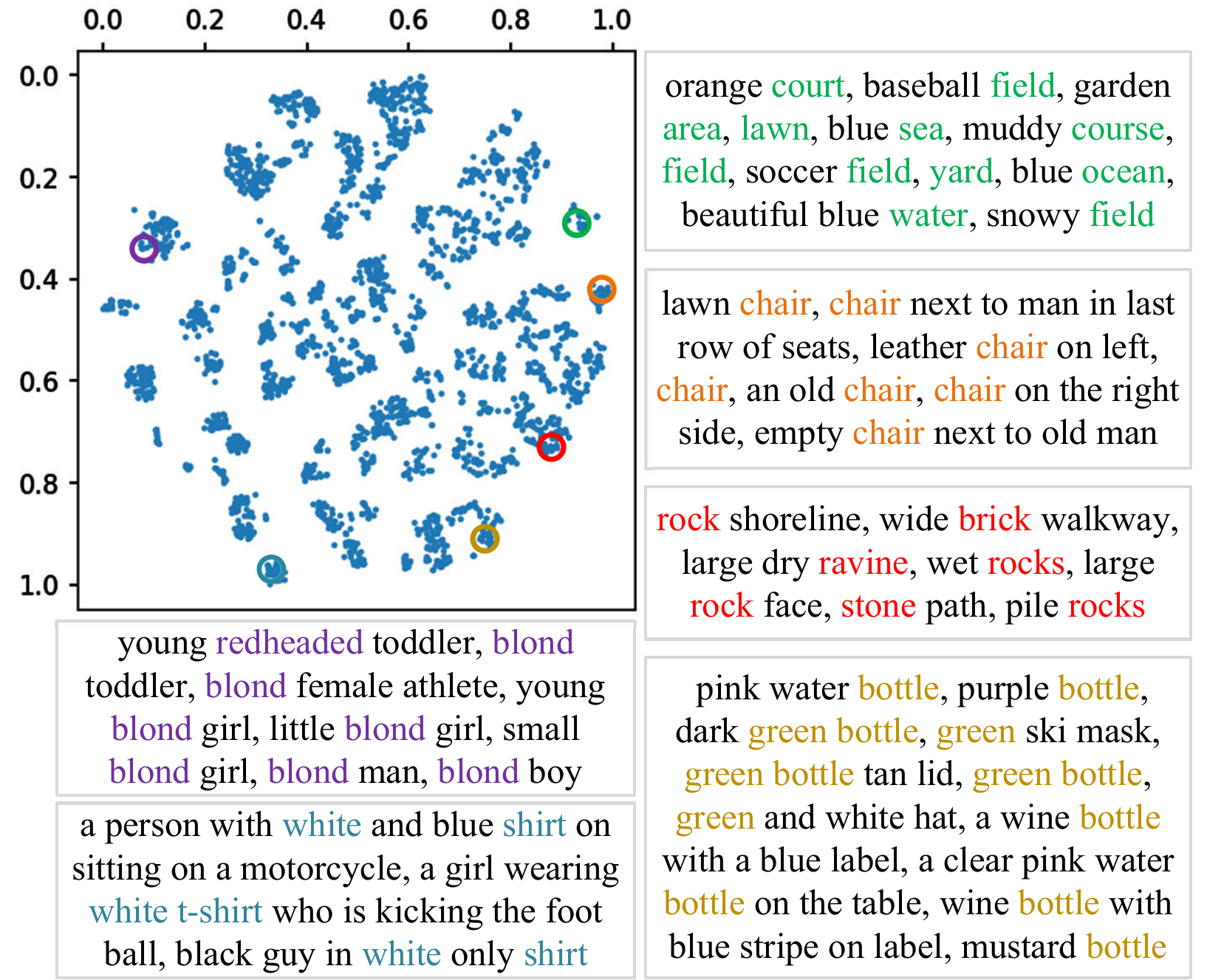}
    \caption{t-SNE visualization of the gate sets, \textit{i.e.} the subnet architectures. Six regions are circled and the corresponding phrases are shown around the figure. Best viewed in color.}
    \label{fig:tsne_text}
\end{figure}

\begin{figure*}[t]
    \centering
    \includegraphics[width=\textwidth]{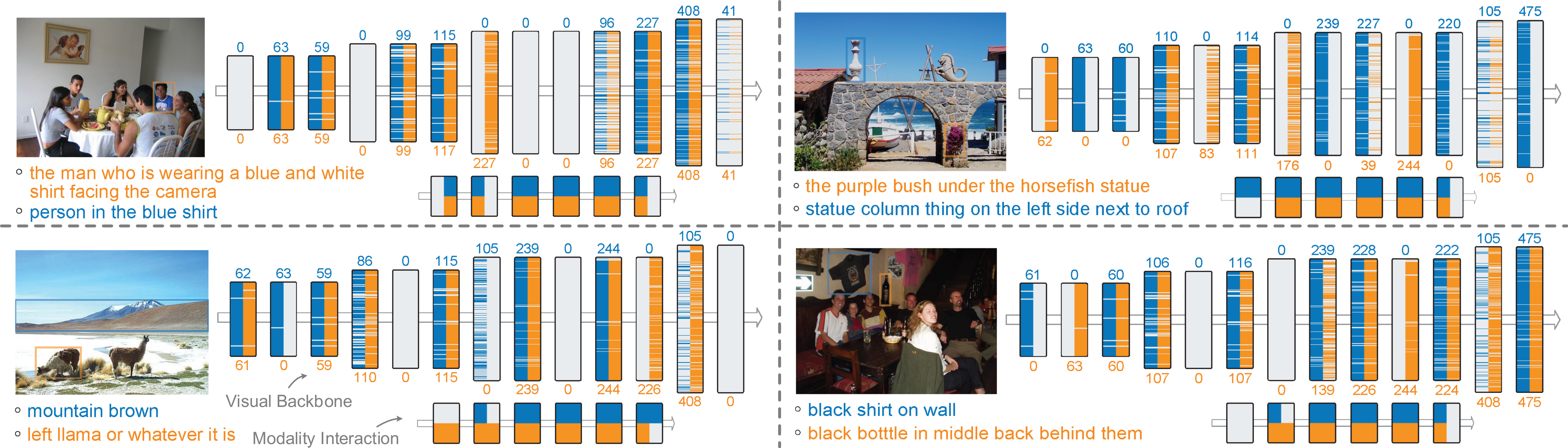}
    \caption{Structures of the expression-adaptive REC subnets. The dynamic visual backbone and modality interaction module are visualized. Selected and unselected filters/layers are marked with color and gray, respectively. 
    The number of selected filters in the four sets of layers with dimensions of 64, 128, 256, and 512 is displayed around the Visual Backbone.
    }
    \label{fig:qualitative_result1}
\end{figure*}

\subsection{Comparison with State-of-the-art Methods}
To estimate the effectiveness of the proposed LADS framework, we conduct quantitative experiments on four widely used datasets, \textit{i.e.} RefCOCO \cite{refcoco}, RefCOCO+ \cite{refcoco}, RefCOCOg \cite{refcocog}, and Referit \cite{referitgame}, and compare results with state-of-the-art methods. The main results are summarized to \cref{tab:main_result}.
These methods can be split into two settings, \textit{i.e.} models trained on each dataset and models pre-trained on large-scale datasets followed by fine-tuning.
The proposed LADS framework achieves the best accuracy and fastest inference speed in both settings.
Compared to the SOTA method Ref-NMS \cite{refnms},  LADS has a better $Prec@0.5$ with +2.15\%/ +2.67\%/ +2.53\% on RefCOCO, +2.91\%/ +3.96\%/ +0.40\%/ on RefCOCO+, and +1.01\%/ +1.04\%/ on RefCOCOg. Compared to the transformer-based method TransVG \cite{transvg}, our method also has better performance with +2.53\%/ +4.00\%/ +0.45\% on RefCOCO, +7.66\%/ +9.49\%/ +4.19\%/ on RefCOCO+, +3.90\%/ +4.22\%/ on RefCOCOg, and +1.32\% on Referit. 
Our model also runs the fastest with only 20ms inference time.

Inspired by recent large-scale pre-training approaches, we also pre-train our method on the large-scale dataset, similar to MDETR \cite{mdetr}, and then fine-tune it on the REC datasets.
Our LADS also outperforms MDETR and has a large accuracy gap compared to the non-pretrained methods, suggesting that large-scale data is necessary for better performance due to the complexity of REC.
Interestingly, our pre-trained model has a 2ms speedup from 20ms to 18ms, which means that better alignment of expressions and subnets can be constrained with large-scale data, resulting in more compact REC subnets.

\subsection{Dynamicity of the REC Subnets}
To verify the dynamicity of the expression-adaptive REC subnets, we count the selection rate of filters and layers on the test set of Referit. 
In addition, the expression-wise numbers of selected layers and filters are also counted. These statistics are illustrated in \cref{fig:filter_layer_disp}.

\cref{fig:filter_layer_disp} (a) shows the dynamicity of the filter selection. 
It can be observed that almost all the filters are highly input-dependent, which means our method can select relevant filters conditioned on various referring expressions. 
There is no never-selected filter on all layers, which means the self-reinforcing problem can be avoided, and thus the model capacity can be preserved.  
\cref{fig:filter_layer_disp} (b) shows the dynamicity of the layer selection. 
It can be seen that both layers in the visual backbone and modality interaction module have high expression adaptability. 
The layers of visual backbone are chosen by nearly half. 
The selection rate of MHSA and FFN in the modality interaction module varies, which means the contributions of the two types of layers are different.

\cref{fig:filter_layer_disp} (c) shows the distribution over the number of selected layers. 
The visual backbone of the dynamic subnets has 8 layers on average and 4 layers at least, which means very shallow visual backbones can be extracted conditioned on the referring expression. 
The modality interaction module of the dynamic subnets has 10 layers on average and 6 layers for some expressions.
\cref{fig:filter_layer_disp} (d) shows the distribution over the number of selected filters.
For simplicity, only 4 layers of filters are visualized. 
It can be observed that the selected filters vary with different referring expressions. 
Moreover, as the number of filters increases, its variance increases, which means the filters of deeper layers in the visual backbone are stronger dynamics.

\begin{figure}
    \centering
    \includegraphics[width=0.45\textwidth]{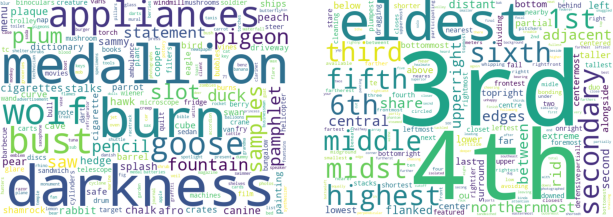}
    \caption{WordCloud of words assigned to the visual backbone (left) and modality interaction module (right). The visual backbone prefers categories and attributes, whereas the modality interaction module prefers context.}
    \label{fig:word_cloud_trans}
\end{figure}
In addition, we also show several specific structures of the expression-adaptive REC subnets in \cref{fig:qualitative_result1}. It can be seen that the LADS framework can extract subnets with different depth/layers and width/channels for various referring expressions. The structures of the REC subnets seem to be similar when the corresponding referring expressions have similar semantic information.

\subsection{Alignment of Expressions and Subnets}
The LADS framework is designed to generate compact REC subnets for any referring expressions, and the architectures of the subnets are desired to align with the expressions, \textit{i.e.} the structures of the subnets corresponding to semantically similar expressions should also be similar.

Since the architectures of subnets can be uniquely determined by the binary gate set $G=\{g_L,g_C\}$, we project the high-dimensional $G$ to the $2$-d coordinates for visualization by using t-SNE. The result is shown in \cref{fig:tsne_text}. Due to the uncountable expressions, there is no obvious clustering, which is different from the classification tasks (\textit{e.g.} ImageNet and CIFAR) with a limited number of categories. 
We visualize the referring expressions corresponding to the architectures in six regions and find that similar architectures do correspond to semantically similar expressions, showing that the expressions and REC subnets are well aligned. 

We also visualize the words assigned to the visual backbone and modality interaction module to perform layer-granularity alignment inspections in \cref{fig:word_cloud_trans}. 
Concretely, we first calculate the word score $s_w\in [0, 1]^{(P+M)}$ for word $w$ by averaging the $\alpha_i$ derived from \cref{equ:token_attn} over all expressions, where $P$ and $M$ represent the number of candidate layers in visual backbone and modality interaction module, respectively. 
Then we normalize $s_w$ to $\hat{s}_w$ using $Softmax$. 
Finally, the scores of word $w$ assigned to the visual backbone and modality interaction module are set to the sum of the first $P$ elements and the last $M$ elements of $\hat{s}_w$, respectively.
Interestingly, the visual backbone prefers categories and attributes  (\textit{e.g.} "medallion" and "darkness"), whereas the modality interaction module prefers context (\textit{e.g.} "3rd", "eldest" and "middle"). 

\section{Conclusions and Future Works}
In this paper, we propose a neat yet efficient framework named Language Adaptive Dynamic Subnets (LADS), which can adaptively extract REC subnets conditioned on the referring expressions. Extensive experiments show that LADS achieves superior inference speed and accuracy compared to state-of-the-art methods, indicating the LADS framework's effectiveness and prospects for more economical and faster REC inference.
For future studies, we think there are two directions worth pursuing, \textit{i.e.}
(1) study the correspondence between language and subnets to achieve more discriminative and accurate subnet selection, and
(2) combine the image-dependent selection mechanism to realize joint selection based on expressions and images.

\section{Acknowledgments}
This work is supported in part by National Key Research and Development Program of China under Grant 2020AAA0107400, Zhejiang Provincial Natural Science Foundation of China under Grant LR19F020004, National Natural Science Foundation of China under Grant U20A20222, National Science Foundation for Distinguished Young Scholars under Grant 62225605, Ant Group, and sponsored by CAAI-HUAWEI MindSpore Open Fund.

\bibliography{aaai23}

\begin{thebibliography}{49}
\providecommand{\natexlab}[1]{#1}

\bibitem[{Bajaj, Wang, and Sigal(2019)}]{bajaj2019g3raphground}
Bajaj, M.; Wang, L.; and Sigal, L. 2019.
\newblock G3raphground: Graph-based language grounding.
\newblock In \emph{Int. Conf. Comput. Vis.}

\bibitem[{Bejnordi, Blankevoort, and Welling(2019)}]{bejnordi2019batch}
Bejnordi, B.~E.; Blankevoort, T.; and Welling, M. 2019.
\newblock Batch-shaping for learning conditional channel gated networks.
\newblock \emph{arXiv preprint arXiv:1907.06627}.

\bibitem[{Cai et~al.(2019)Cai, Gan, Wang, Zhang, and Han}]{ofa}
Cai, H.; Gan, C.; Wang, T.; Zhang, Z.; and Han, S. 2019.
\newblock Once-for-All: Train One Network and Specialize it for Efficient
  Deployment.
\newblock In \emph{Int. Conf. Learn. Represent.}

\bibitem[{Carion et~al.(2020)Carion, Massa, Synnaeve, Usunier, Kirillov, and
  Zagoruyko}]{carion2020end}
Carion, N.; Massa, F.; Synnaeve, G.; Usunier, N.; Kirillov, A.; and Zagoruyko,
  S. 2020.
\newblock End-to-end object detection with transformers.
\newblock In \emph{Eur. Conf. Comput. Vis.}, 213--229.

\bibitem[{Chen et~al.(2019{\natexlab{a}})Chen, Zhu, Li, and
  Zhao}]{chen2019self}
Chen, J.; Zhu, Z.; Li, C.; and Zhao, Y. 2019{\natexlab{a}}.
\newblock Self-adaptive network pruning.
\newblock In \emph{Adv. Neural Inform. Process. Syst.}, 175--186.

\bibitem[{Chen et~al.(2021)Chen, Ma, Xiao, Zhang, and Chang}]{refnms}
Chen, L.; Ma, W.; Xiao, J.; Zhang, H.; and Chang, S.-F. 2021.
\newblock Ref-NMS: breaking proposal bottlenecks in two-stage referring
  expression grounding.
\newblock In \emph{AAAI}.

\bibitem[{Chen et~al.(2020)Chen, Li, Yu, El~Kholy, Ahmed, Gan, Cheng, and
  Liu}]{uniter}
Chen, Y.-C.; Li, L.; Yu, L.; El~Kholy, A.; Ahmed, F.; Gan, Z.; Cheng, Y.; and
  Liu, J. 2020.
\newblock Uniter: Universal image-text representation learning.
\newblock In \emph{Eur. Conf. Comput. Vis.}, 104--120.

\bibitem[{Chen et~al.(2019{\natexlab{b}})Chen, Li, Bengio, and
  Si}]{chen2019you}
Chen, Z.; Li, Y.; Bengio, S.; and Si, S. 2019{\natexlab{b}}.
\newblock You look twice: Gaternet for dynamic filter selection in cnns.
\newblock In \emph{IEEE Conf. Comput. Vis. Pattern Recog.}, 9172--9180.

\bibitem[{Deng et~al.(2021)Deng, Yang, Chen, Zhou, and Li}]{transvg}
Deng, J.; Yang, Z.; Chen, T.; Zhou, W.; and Li, H. 2021.
\newblock Transvg: End-to-end visual grounding with transformers.
\newblock In \emph{Int. Conf. Comput. Vis.}

\bibitem[{Devlin et~al.(2018)Devlin, Chang, Lee, and Toutanova}]{bert}
Devlin, J.; Chang, M.-W.; Lee, K.; and Toutanova, K. 2018.
\newblock Bert: Pre-training of deep bidirectional transformers for language
  understanding.
\newblock \emph{arXiv preprint arXiv:1810.04805}.

\bibitem[{Escalante et~al.(2010)Escalante, Hern{\'a}ndez, Gonzalez,
  L{\'o}pez-L{\'o}pez, Montes, Morales, Sucar, Villasenor, and
  Grubinger}]{saiapr12}
Escalante, H.~J.; Hern{\'a}ndez, C.~A.; Gonzalez, J.~A.; L{\'o}pez-L{\'o}pez,
  A.; Montes, M.; Morales, E.~F.; Sucar, L.~E.; Villasenor, L.; and Grubinger,
  M. 2010.
\newblock The segmented and annotated IAPR TC-12 benchmark.
\newblock \emph{CVIU}.

\bibitem[{Gan et~al.(2020)Gan, Chen, Li, Zhu, Cheng, and Liu}]{villa}
Gan, Z.; Chen, Y.-C.; Li, L.; Zhu, C.; Cheng, Y.; and Liu, J. 2020.
\newblock Large-scale adversarial training for vision-and-language
  representation learning.
\newblock \emph{Advances in Neural Information Processing Systems}, 33:
  6616--6628.

\bibitem[{Gao et~al.(2018)Gao, Zhao, Dudziak, Mullins, and Xu}]{gao2018dynamic}
Gao, X.; Zhao, Y.; Dudziak, {\L}.; Mullins, R.; and Xu, C.-z. 2018.
\newblock Dynamic channel pruning: Feature boosting and suppression.
\newblock \emph{arXiv preprint arXiv:1810.05331}.

\bibitem[{Glorot and Bengio(2010)}]{glorot2010understanding}
Glorot, X.; and Bengio, Y. 2010.
\newblock Understanding the difficulty of training deep feedforward neural
  networks.
\newblock In \emph{Proceedings of the thirteenth international conference on
  artificial intelligence and statistics}.

\bibitem[{He et~al.(2017)He, Gkioxari, Doll{\'a}r, and Girshick}]{he2017mask}
He, K.; Gkioxari, G.; Doll{\'a}r, P.; and Girshick, R. 2017.
\newblock Mask r-cnn.
\newblock In \emph{IEEE Conf. Comput. Vis. Pattern Recog.}

\bibitem[{He et~al.(2016)He, Zhang, Ren, and Sun}]{resnet}
He, K.; Zhang, X.; Ren, S.; and Sun, J. 2016.
\newblock Deep residual learning for image recognition.
\newblock In \emph{IEEE Conf. Comput. Vis. Pattern Recog.}

\bibitem[{Herrmann, Bowen, and Zabih(2020)}]{herrmann2020channel}
Herrmann, C.; Bowen, R.~S.; and Zabih, R. 2020.
\newblock Channel selection using gumbel softmax.
\newblock In \emph{Eur. Conf. Comput. Vis.}, 241--257.

\bibitem[{Hong et~al.(2019)Hong, Liu, Mo, He, and Zhang}]{hong2019learning}
Hong, R.; Liu, D.; Mo, X.; He, X.; and Zhang, H. 2019.
\newblock Learning to compose and reason with language tree structures for
  visual grounding.
\newblock \emph{IEEE Trans. Pattern Anal. Mach. Intell.}

\bibitem[{Jang, Gu, and Poole(2016)}]{jang2016categorical}
Jang, E.; Gu, S.; and Poole, B. 2016.
\newblock Categorical reparameterization with gumbel-softmax.
\newblock \emph{arXiv preprint arXiv:1611.01144}.

\bibitem[{Kaiser and Bengio(2018)}]{kaiser2018discrete}
Kaiser, {\L}.; and Bengio, S. 2018.
\newblock Discrete autoencoders for sequence models.
\newblock \emph{arXiv preprint arXiv:1801.09797}.

\bibitem[{Kamath et~al.(2021)Kamath, Singh, LeCun, Synnaeve, Misra, and
  Carion}]{mdetr}
Kamath, A.; Singh, M.; LeCun, Y.; Synnaeve, G.; Misra, I.; and Carion, N. 2021.
\newblock MDETR-modulated detection for end-to-end multi-modal understanding.
\newblock In \emph{Int. Conf. Comput. Vis.}, 1780--1790.

\bibitem[{Kazemzadeh et~al.(2014)Kazemzadeh, Ordonez, Matten, and
  Berg}]{referitgame}
Kazemzadeh, S.; Ordonez, V.; Matten, M.; and Berg, T. 2014.
\newblock Referitgame: Referring to objects in photographs of natural scenes.
\newblock In \emph{EMNLP}.

\bibitem[{Krishna et~al.(2017)Krishna, Zhu, Groth, Johnson, Hata, Kravitz,
  Chen, Kalantidis, Li, Shamma et~al.}]{vg}
Krishna, R.; Zhu, Y.; Groth, O.; Johnson, J.; Hata, K.; Kravitz, J.; Chen, S.;
  Kalantidis, Y.; Li, L.-J.; Shamma, D.~A.; et~al. 2017.
\newblock Visual genome: Connecting language and vision using crowdsourced
  dense image annotations.
\newblock \emph{Int. J. Comput. Vis.}

\bibitem[{Li et~al.(2018)Li, Fu, Yu, Mei, and Luo}]{li2018tell}
Li, Q.; Fu, J.; Yu, D.; Mei, T.; and Luo, J. 2018.
\newblock Tell-and-answer: Towards explainable visual question answering using
  attributes and captions.
\newblock \emph{EMNLP}.

\bibitem[{Liao et~al.(2020)Liao, Liu, Li, Wang, Chen, Qian, and Li}]{rccf}
Liao, Y.; Liu, S.; Li, G.; Wang, F.; Chen, Y.; Qian, C.; and Li, B. 2020.
\newblock A real-time cross-modality correlation filtering method for referring
  expression comprehension.
\newblock In \emph{IEEE Conf. Comput. Vis. Pattern Recog.}

\bibitem[{Lin et~al.(2014)Lin, Maire, Belongie, Hays, Perona, Ramanan,
  Doll{\'a}r, and Zitnick}]{mscoco}
Lin, T.-Y.; Maire, M.; Belongie, S.; Hays, J.; Perona, P.; Ramanan, D.;
  Doll{\'a}r, P.; and Zitnick, C.~L. 2014.
\newblock Microsoft coco: Common objects in context.
\newblock In \emph{Eur. Conf. Comput. Vis.}

\bibitem[{Liu et~al.(2019)Liu, Zhang, Wu, and Zha}]{liu2019learning}
Liu, D.; Zhang, H.; Wu, F.; and Zha, Z.-J. 2019.
\newblock Learning to assemble neural module tree networks for visual
  grounding.
\newblock In \emph{Int. Conf. Comput. Vis.}

\bibitem[{Liu, Wang, and Yang(2017)}]{liu2017referring}
Liu, J.; Wang, L.; and Yang, M.-H. 2017.
\newblock Referring expression generation and comprehension via attributes.
\newblock In \emph{Int. Conf. Comput. Vis.}

\bibitem[{Loshchilov and Hutter(2020)}]{adamw}
Loshchilov, I.; and Hutter, F. 2020.
\newblock Decoupled weight decay regularization.
\newblock In \emph{Int. Conf. Learn. Represent.}

\bibitem[{Lu et~al.(2019)Lu, Batra, Parikh, and Lee}]{vilbert}
Lu, J.; Batra, D.; Parikh, D.; and Lee, S. 2019.
\newblock Vilbert: Pretraining task-agnostic visiolinguistic representations
  for vision-and-language tasks.
\newblock \emph{Advances in neural information processing systems}, 32.

\bibitem[{Luo et~al.(2020)Luo, Zhou, Sun, Cao, Wu, Deng, and Ji}]{mcn}
Luo, G.; Zhou, Y.; Sun, X.; Cao, L.; Wu, C.; Deng, C.; and Ji, R. 2020.
\newblock Multi-task collaborative network for joint referring expression
  comprehension and segmentation.
\newblock In \emph{IEEE Conf. Comput. Vis. Pattern Recog.}

\bibitem[{Maddison, Mnih, and Teh(2016)}]{maddison2016concrete}
Maddison, C.~J.; Mnih, A.; and Teh, Y.~W. 2016.
\newblock The concrete distribution: A continuous relaxation of discrete random
  variables.
\newblock \emph{arXiv preprint arXiv:1611.00712}.

\bibitem[{Mao et~al.(2016)Mao, Huang, Toshev, Camburu, Yuille, and
  Murphy}]{refcocog}
Mao, J.; Huang, J.; Toshev, A.; Camburu, O.; Yuille, A.~L.; and Murphy, K.
  2016.
\newblock Generation and comprehension of unambiguous object descriptions.
\newblock In \emph{IEEE Conf. Comput. Vis. Pattern Recog.}

\bibitem[{Nagaraja, Morariu, and Davis(2016)}]{umd}
Nagaraja, V.~K.; Morariu, V.~I.; and Davis, L.~S. 2016.
\newblock Modeling context between objects for referring expression
  understanding.
\newblock In \emph{Eur. Conf. Comput. Vis.}

\bibitem[{Plummer et~al.(2015)Plummer, Wang, Cervantes, Caicedo, Hockenmaier,
  and Lazebnik}]{flickr30k}
Plummer, B.~A.; Wang, L.; Cervantes, C.~M.; Caicedo, J.~C.; Hockenmaier, J.;
  and Lazebnik, S. 2015.
\newblock Flickr30k entities: Collecting region-to-phrase correspondences for
  richer image-to-sentence models.
\newblock In \emph{Int. Conf. Comput. Vis.}

\bibitem[{Ren et~al.(2015)Ren, He, Girshick, and Sun}]{ren2015faster}
Ren, S.; He, K.; Girshick, R.; and Sun, J. 2015.
\newblock Faster r-cnn: Towards real-time object detection with region proposal
  networks.
\newblock \emph{Adv. Neural Inform. Process. Syst.}

\bibitem[{Salvador et~al.(2016)Salvador, Gir{\'o}-i Nieto, Marqu{\'e}s, and
  Satoh}]{salvador2016faster}
Salvador, A.; Gir{\'o}-i Nieto, X.; Marqu{\'e}s, F.; and Satoh, S. 2016.
\newblock Faster r-cnn features for instance search.
\newblock In \emph{IEEE Conf. Comput. Vis. Pattern Recog.}

\bibitem[{Shazeer et~al.(2017)Shazeer, Mirhoseini, Maziarz, Davis, Le, Hinton,
  and Dean}]{shazeer2017outrageously}
Shazeer, N.; Mirhoseini, A.; Maziarz, K.; Davis, A.; Le, Q.; Hinton, G.; and
  Dean, J. 2017.
\newblock Outrageously large neural networks: The sparsely-gated
  mixture-of-experts layer.
\newblock \emph{arXiv preprint arXiv:1701.06538}.

\bibitem[{Veit and Belongie(2018)}]{veit2018convolutional}
Veit, A.; and Belongie, S. 2018.
\newblock Convolutional networks with adaptive inference graphs.
\newblock In \emph{Eur. Conf. Comput. Vis.}, 3--18.

\bibitem[{Wang et~al.(2019)Wang, Wu, Cao, Shen, Gao, and
  Hengel}]{wang2019neighbourhood}
Wang, P.; Wu, Q.; Cao, J.; Shen, C.; Gao, L.; and Hengel, A. v.~d. 2019.
\newblock Neighbourhood watch: Referring expression comprehension via
  language-guided graph attention networks.
\newblock In \emph{IEEE Conf. Comput. Vis. Pattern Recog.}

\bibitem[{Wang et~al.(2018)Wang, Yu, Dou, Darrell, and
  Gonzalez}]{wang2018skipnet}
Wang, X.; Yu, F.; Dou, Z.-Y.; Darrell, T.; and Gonzalez, J.~E. 2018.
\newblock Skipnet: Learning dynamic routing in convolutional networks.
\newblock In \emph{Eur. Conf. Comput. Vis.}, 409--424.

\bibitem[{Wolf et~al.(2020)Wolf, Debut, Sanh, Chaumond, Delangue, Moi, Cistac,
  Rault, Louf, Funtowicz et~al.}]{huggingface}
Wolf, T.; Debut, L.; Sanh, V.; Chaumond, J.; Delangue, C.; Moi, A.; Cistac, P.;
  Rault, T.; Louf, R.; Funtowicz, M.; et~al. 2020.
\newblock Transformers: State-of-the-art natural language processing.
\newblock In \emph{EMNLP}.

\bibitem[{Xia et~al.(2021)Xia, Yin, Dai, and Jha}]{xia2021fully}
Xia, W.; Yin, H.; Dai, X.; and Jha, N.~K. 2021.
\newblock Fully dynamic inference with deep neural networks.
\newblock \emph{IEEE Transactions on Emerging Topics in Computing}.

\bibitem[{Yang, Li, and Yu(2019)}]{yang2019cross}
Yang, S.; Li, G.; and Yu, Y. 2019.
\newblock Cross-modal relationship inference for grounding referring
  expressions.
\newblock In \emph{IEEE Conf. Comput. Vis. Pattern Recog.}

\bibitem[{Yang et~al.(2020)Yang, Chen, Wang, and Luo}]{resc}
Yang, Z.; Chen, T.; Wang, L.; and Luo, J. 2020.
\newblock Improving one-stage visual grounding by recursive sub-query
  construction.
\newblock In \emph{Eur. Conf. Comput. Vis.}

\bibitem[{Yang et~al.(2019)Yang, Gong, Wang, Huang, Yu, and Luo}]{faoa}
Yang, Z.; Gong, B.; Wang, L.; Huang, W.; Yu, D.; and Luo, J. 2019.
\newblock A fast and accurate one-stage approach to visual grounding.
\newblock In \emph{Int. Conf. Comput. Vis.}, 4683--4693.

\bibitem[{Yu et~al.(2018)Yu, Lin, Shen, Yang, Lu, Bansal, and Berg}]{mattnet}
Yu, L.; Lin, Z.; Shen, X.; Yang, J.; Lu, X.; Bansal, M.; and Berg, T.~L. 2018.
\newblock Mattnet: Modular attention network for referring expression
  comprehension.
\newblock In \emph{IEEE Conf. Comput. Vis. Pattern Recog.}

\bibitem[{Yu et~al.(2016)Yu, Poirson, Yang, Berg, and Berg}]{refcoco}
Yu, L.; Poirson, P.; Yang, S.; Berg, A.~C.; and Berg, T.~L. 2016.
\newblock Modeling context in referring expressions.
\newblock In \emph{Eur. Conf. Comput. Vis.}

\bibitem[{Zhou et~al.(2021)Zhou, Ji, Luo, Sun, Su, Ding, Lin, and
  Tian}]{zhou2021real}
Zhou, Y.; Ji, R.; Luo, G.; Sun, X.; Su, J.; Ding, X.; Lin, C.-W.; and Tian, Q.
  2021.
\newblock A real-time global inference network for one-stage referring
  expression comprehension.
\newblock \emph{IEEE Transactions on Neural Networks and Learning Systems}.

\end{thebibliography}

\end{document}